\documentclass[letterpaper,10pt,conference]{ieeeconf}

\IEEEoverridecommandlockouts %
\usepackage{graphics} %
\usepackage{times} %
\usepackage{amsmath} %
\usepackage{amssymb}  %
\usepackage{bm}
\usepackage[export]{adjustbox} %
\usepackage{algorithm, algpseudocode}

\makeatletter
\newcommand\fs@spaceruled{\def\@fs@cfont{\bfseries}\let\@fs@capt\floatc@ruled
  \def\@fs@pre{\vspace{4pt}\hrule height.8pt depth0pt \kern2pt}%
  \def\@fs@post{\kern2pt\hrule\relax}%
  \def\@fs@mid{\kern2pt\hrule\kern2pt}%
  \let\@fs@iftopcapt\iftrue}
\makeatother

\floatstyle{spaceruled}%
\restylefloat{algorithm}

\usepackage{caption}
\usepackage{subcaption}
\usepackage{mathtools}

\usepackage[noadjust]{cite} 
\usepackage[bookmarks=true,hidelinks]{hyperref}

\usepackage[binary-units=true]{siunitx}
\usepackage{cleveref}

\usepackage[latin1]{inputenc}
\usepackage[english]{babel}
\usepackage{todonotes}

\usepackage{booktabs}
\usepackage{siunitx}

\newcommand{\dd}{\text{d}}

\newcommand{\bu}{\bm{u}}
\newcommand{\bx}{\bm{x}}
\newcommand{\bfun}{\bm{f}}

\newcommand{\delx}{\delta \bx}
\newcommand{\delu}{\delta \bu}

\newcommand{\eref}[1]{Eq.~(\ref{#1})}

\newcommand\copyrighttext{%
  \footnotesize \textcopyright This work has been submitted to the IEEE for possible publication. Copyright may be transferred without notice, after which this version may no longer be accessible.}
\newcommand\copyrightnotice{%
\begin{tikzpicture}[remember picture,overlay]
\node[anchor=south,yshift=10pt] at (current page.south) {\fbox{\parbox{\dimexpr\textwidth-\fboxsep-\fboxrule\relax}{\copyrighttext}}};
\end{tikzpicture}%
}

\DeclareMathOperator*{\argmin}{arg\,min} %

\title{\LARGE \bf
    Differentiable Optimal Control via Differential Dynamic Programming
}

\author{Traiko Dinev$^1$\quad Carlos Mastalli$^{2,3}$\quad Vladimir Ivan$^1$\quad Steve Tonneau$^1$\quad Sethu Vijayakumar$^1$%
\thanks{$^1$School of Informatics, The University of Edinburgh, Edinburgh, UK.}
\thanks{$^{2}$Institute of Sensors, Signals and Systems, School of Engineering and Physical Sciences, Heriot-Watt University, Edinburgh EH14 4AS, UK.}%
\thanks{$^{3}$National Robotarium, Edinburgh, UK}
\thanks{This research is supported by the EPSRC Centre for Doctoral Training in Robotics and Autonomous Systems (EP/L016834/1).}
}

\begin{document}

\maketitle
\copyrightnotice
\thispagestyle{empty}
\pagestyle{empty}

\begin{abstract}
Robot design optimization, imitation learning and system identification share a common problem which requires optimization over robot or task parameters at the same time as optimizing the robot motion. To solve these problems, we can use \textit{differentiable optimal control} for which the gradients of the robot's motion with respect to the parameters are required. We propose a method to efficiently compute these gradients analytically via the \textit{differential dynamic programming} (DDP) algorithm using sensitivity analysis (SA). We show that we must include second-order dynamics terms when computing the gradients. However, we do not need to include them when computing the motion. We validate our approach on the pendulum and double pendulum systems. Furthermore, we compare against using the derivatives of the iterative linear quadratic regulator (iLQR), which ignores these second-order terms everywhere, on a co-design task for the Kinova arm, where we optimize the link lengths of the robot for a target reaching task. We show that optimizing using iLQR gradients diverges as ignoring the second-order dynamics affects the computation of the derivatives. Instead, optimizing using DDP gradients converges to the same optimum for a range of initial designs  allowing our formulation to scale to complex systems.
\end{abstract}

\section{Introduction}
Optimal control (OC) is a popular tool for planning dynamic motions for robots (\cite{mastalli2020direct, mastalli_crocoddyl_2020, ferrolho2021inverse}). To specify such kind of problems, we are required to provide (i) a set of constraints that ensure a physically-realizable motion and (ii) a set of cost functions that describe the desired task. These constraints and costs terms can be described through a set of hyper-parameters. Those can include cost weights, as well as properties of the robot, such as masses and link lengths, that define its geometry and dynamics.

We can also use optimal control to optimize these cost function weights and robot physical properties. We consider two applications of this. The first is imitation learning, where we optimize the cost function weights to imitate a teacher's cost function. The second is co-design, where we optimize the robot design for a given task. We address both by using gradient optimization, for which we require the derivative of the optimal control problem. We refer to this technique as \textit{differentiable optimal control}.

\begin{figure}[t]
    \centering
    \includegraphics[width=0.225 \textwidth, trim={25cm 5cm 20cm 5cm},clip]{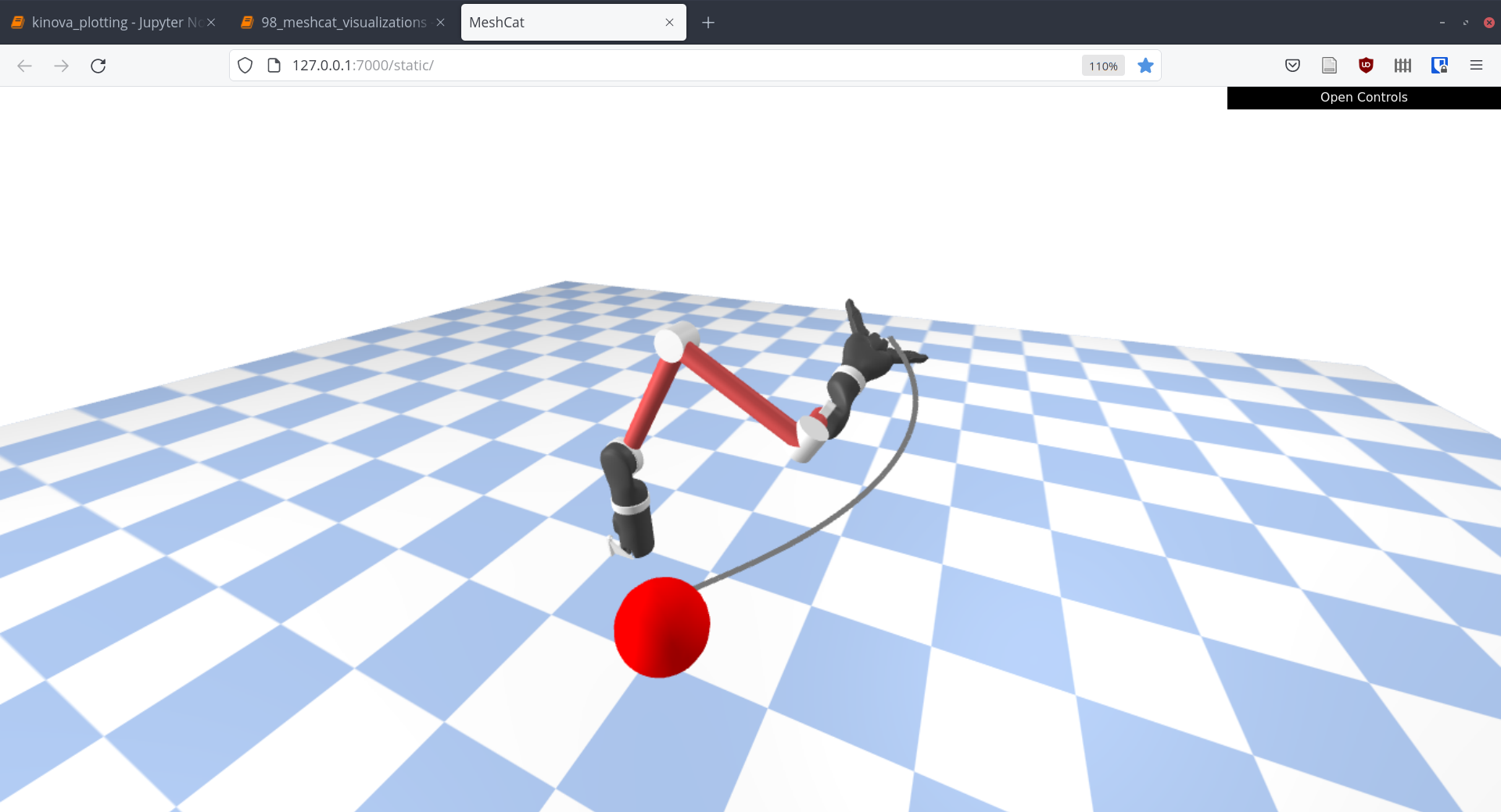}
    \includegraphics[width=0.225 \textwidth, trim={25cm 5cm 20cm 5cm},clip]{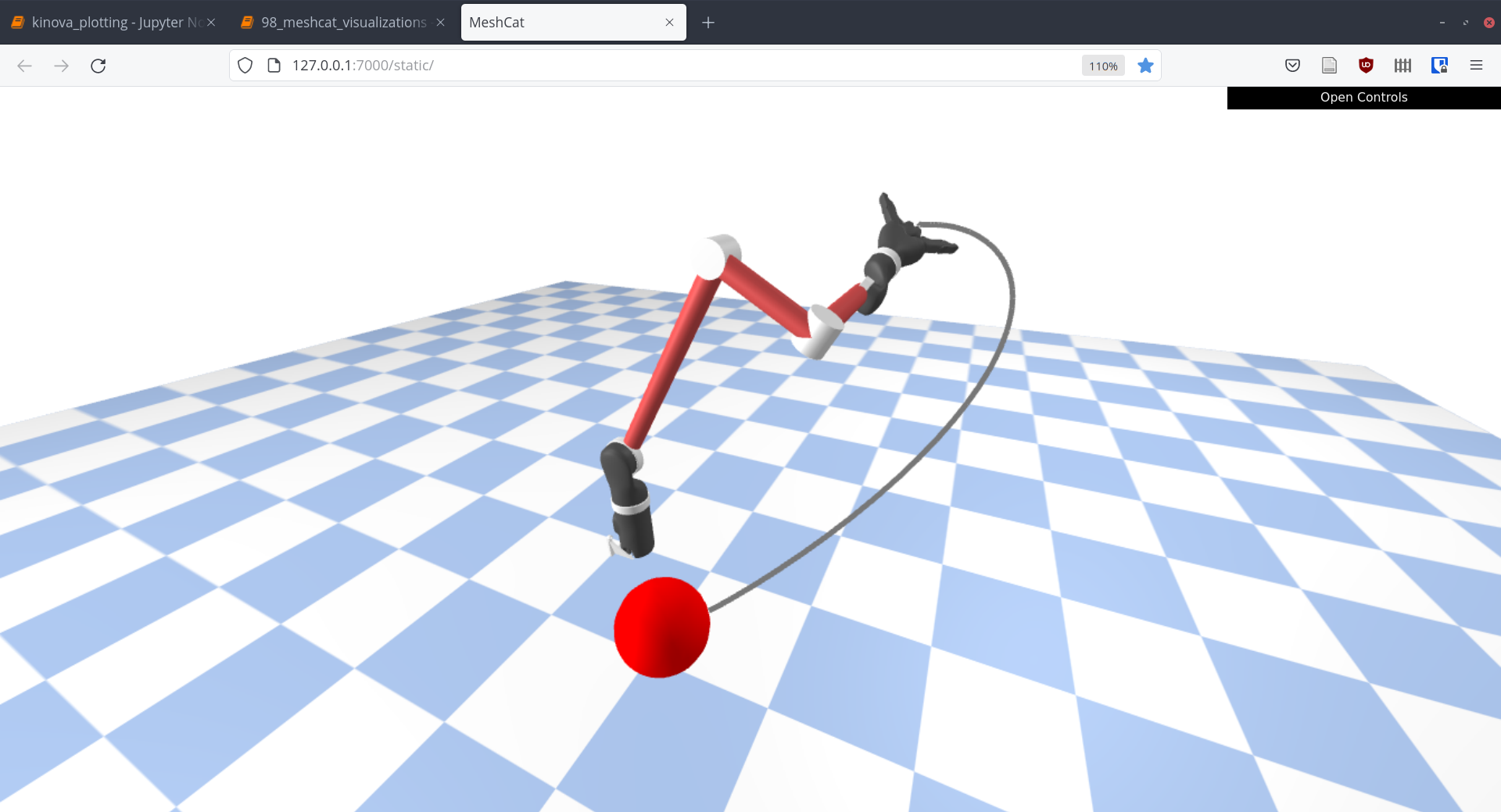}
    \caption{Optimized robot design for minimum joint velocity (left) and initial design (right) for the Kinova arm. The optimized robot has the end-effector at the same z-plane as the target (in red), thus needing only the base to move, which minimizes joint velocity. Please find the accompanying video at  \url{https://youtu.be/riXP9k2PUVs}}
    \label{fig:poster}
\end{figure}

Differentiable optimal control can be done through numerical and automatic differentiation~\cite{dinev2021co}, or sensitivity analysis (SA)~\cite{gould2016differentiating}. Numerical differentiation suffers from low numerical accuracy and automatic differentiation (AD) involves the unrolling of the entire optimization loop, which suffers from slow computational times and requires the optimization to be written in such a way so that AD tools can differentiate through it~\cite{amos_diffmpc_2018}. Sensitivity analysis (SA) offers better computational speed, often requiring a single iteration through time. Moreover, using SA we can build on top of existing solvers without requiring AD support.

\subsection{Contributions}
In this work we present a sensitivity analysis approach for differentiating optimal control problems via differential dynamic programming (DDP), which is a second-order approach to solving OC problems~\cite{koenemann2015whole,neunert-ral18}. In the literature a related 1.5-order approach known as the iterative linear quadratic regulator~\cite{li_iterative_2004} (iLQR) is often preferred, as it does not use second-order derivative tensors. We show that these second-order terms are actually necessary for exact differentiable optimal control algorithms. However, they are only needed during the computation of the derivatives, and not when solving the optimal control problem itself. This allows us to use any method (including either iLQR or DDP) for solving optimal control problems, which we show by using the optimal control library \textsc{Crocoddyl}~\cite{mastalli_crocoddyl_2020}.

Our work provides a complete derivation of the analytical derivatives of nonlinear optimal control problems via DDP. In particular, we have the following technical contribution:

\begin{enumerate}
    \item An approach to efficiently compute analytical derivatives using second-order DDP terms only once in a backward pass (but not while solving the OC problem itself) that avoids the unrolling used in AD.
    \item A set of experiments validating the numerical correctness of gradients computed by our approach on the pendulum and double pendulum systems.
    \item A co-design experiment and methodology that shows how to scale our method to the Kinova robot arm using an existing optimal control solver (\textsc{crocoddyl}~\cite{mastalli_crocoddyl_2020}) and differentiate through it. 
\end{enumerate}

\subsection{Related Work}
Recently, there has been a large interest in applying sensitivity analysis to optimal control (OC) problems. Two fields have contributed to this development. The first is concurrent design (co-design), which aims to optimize the design of the robot. The second is machine learning, where system identification and cost learning are the main goals.

One of the first works to differentiate an OC problem in robotics is \cite{ha_computational_2018}, where the method is based on an implicit relationship between motion and design. It also linearizes the constraints, in the optimization problem, to obtain the required gradients. However, this approach still requires the use of numerical approximation for the computation of the gradients and the linearization leads to errors.

Building on this work, \cite{desai_interactive_2018,geilinger_skaterbots:_2018,geilinger_computational_2020} propose a technique to differentiate through an unconstrained OC problem. This technique allows them to formulate the nonlinear relationship between design and motion easily. However, all of these approaches involve inverting the so-called Karush-Kuhn-Tucker (KKT) matrix directly, which is numerically expensive. Moreover, this approach requires the use of unconstrained optimization, which does not guarantee dynamic feasibility. In this letter we consider constrained optimization and use a linear-quadratic regulator to efficiently compute the derivatives without inverting the KKT matrix directly.

Recently, a new body of work proposes an efficient method for differentiating LQR and iLQR problems by employing sensitivity analysis on their KKT conditions~\cite{amos_optnet_2019,amos_diffmpc_2018}. This technique exploits the structure of the KKT matrix and reduces the computation needed to invert/factorize it. However, as we will show, using this approach leads to errors on the double pendulum and Kinova arm due to the iLQR approximation. Indeed, the second-order dynamics derivatives included in DDP but excluded in iLQR are necessary to compute the derivatives exactly.

\section{Background}
We first define the optimal control problem and showing how to solve it. Then we show how to compute its derivative. %

\subsection{Nonlinear Optimal Control Problem} \label{sec:shooting_oc}
Consider the nonlinear optimal control (OC) problem parameterized by the vector $\bm{\theta}$:
\begin{align} \label{eq:oc_cost}
    \min_{X,U} J_\textsc{OC}(X, U; \bm{\theta}) &= h(\bx_T; \bm{\theta}) + \sum_{t = 1}^{T - 1} \ell(\bx_t, \bu_t; \bm{\theta}) \nonumber\\
    \text{s.t}\ \bm{x}_{t + 1} &= \bfun(\bx_t, \bu_t; \bm{\theta}) \\\nonumber
        \bm{x}_1 &= \bm{\bar{x}}_1,
\end{align}
where $\bx = [\bm{q}, \dot{\bm{q}}]$ is the state of the system, composed of its generalized coordinates and velocities ($\bm{q}$, $\dot{\bm{q}}$). The nonlinear system evolves in time according to its dynamics $\bfun(\bx_t, \bu_t; \bm{\theta})$, $X = \{ \bx_1, \cdots, \bx_T \}$ is the set of states where $T$ is number of states discretized in time, $U = \{ \bu_1, \cdots, \bu_{T - 1} \}$ is the set of controls, $\bm{\bar{x}}_1$ defines the starting state, and $\bx_T$ is the terminal state. $\{X, U\}$ is known as the system trajectory.

Subject to the dynamics constraints, the OC problem minimizes a set of nonlinear running costs $\ell(\bx_t, \bu_t; \bm{\theta})$, and a terminal cost $h(\bx_T; \bm{\theta})$. These costs allow us to define the goal of our OC problem. For instance, we could set-up $h(\bx_T; \bm{\theta}) = (\bx_T - \bm{x}^*)^\top \bm{Q}_f (\bx_T - \bm{x}^*)$ and $\ell(\bx_t, \bu_t;\bm{\theta}) = \bu_t^\top \mathbf{R} \bu_t$, where $\bm{x}^*$ is the goal state and the cost weights $\bm{Q}_f$, $\bm{R}$ penalize distance from the goal state and large control inputs, respectively. Then the OC problem finds a trajectory that reaches the goal state with minimum control input. In this letter we optimize some of the parameters of the OC problem, which can be the weights of the cost functions (e.g. $\bm{\theta} = \{\bm{Q}_f,\bm{R}\}$) when performing imitation learning or dynamic properties of the robot when optimizing its design.

\subsection{Solving the Optimal Control Problem}\label{sec:ddp}
We can solve the OC problem for a fixed $\bm{\theta}$ using DDP~\cite{mayne1966ddp} or iLQR~\cite{li_iterative_2004}. In the following, we omit $\bm{\theta}$ for clarity. We begin by defining the Value function as the optimal cost-to-go starting at time $t$:
\begin{equation}
    \mathcal{V}(\bx, t) \triangleq \min_{\bu_t, \dots, \bu_{T - 1}} h(\bx_T) + \sum_{i = t}^{T - 1} \ell(\bu_i, \bx_i).
\end{equation}

We then define the action-value function (Q-function) as the optimal cost-to-go plus the current state-action cost:
\begin{equation} \label{eq:oc_q}
    Q(\bx, \bu, t) = \ell(\bx, \bu) + \mathcal{V}(\bfun(\bx, \bu), t + 1).
\end{equation}
DDP and iLQR then iterate between (i) minimizing the quadratic approximation of the Q-function in a backward pass and (ii) integrating the system dynamics in a forward pass. They differ in the approximation used as we describe next.

In the backward pass we approximate the Q-function by a second order Taylor series around a reference trajectory $U^r = \{ \bu^r_1, \dots, \bu^r_{T - 1} \}$ and $X^r = \{ \bx^r_1, \dots, \bx^r_T \}$:
\begin{gather} \label{eq:q-expansion}
    \begin{split}
        Q(\bx, \bu) = Q(\bx^r, \bu^r) + 
        \bm{Q_x} \delx + \mathbf{Q_u} \delu +\\ \frac{1}{2} \delx^\top \bm{Q_{xx}} \delx + 
            \frac{1}{2} \delu^\top \bm{Q_{uu}} \delu + \delu^\top \bm{Q_{ux}} \delx.
    \end{split}
\end{gather}
This approximation is done in the tangential space (i.e., $\delta$-space), such that $\delx = \bx - \bx^r$ and $\delu = \bu - \bu^r$. We then give the derivatives of the Q-function, where $\mathcal{V}'$ is the value function at the next state:
\begin{align} \label{eq:ddp_backward_pass}
    \bm{Q_x} &= \bm{\ell_x} + \bfun_{\bm{x}}^\top \mathcal{V}'_{\bm{x}}, \nonumber\\
    \bm{Q_u} &= \bm{\ell_u} + \bfun_{\bm{u}}^\top \mathcal{V}'_{\bm{x}}, \nonumber\\
    \bm{Q_{xx}} &= \bm{\ell_{xx}} + \bfun_{\bm{x}}^\top \mathcal{V}'_{\bm{xx}} \bfun_{\bm{x}}^\top + \underbrace{\mathcal{V}'_{\bm{x}} \cdot \bfun_{\bm{xx}}}, \nonumber\\
    \bm{Q_{uu}} &= \bm{\ell_{uu}} + \bfun_{\bm{u}}^\top \mathcal{V}'_{\bm{xx}} \bfun_{\bm{u}}^\top + \underbrace{\mathcal{V}'_{\bm{x}} \cdot \bfun_{\bm{uu}}}, \nonumber\\
    \bm{Q_{ux}} &= \bm{\ell_{ux}} + \bfun_{\bm{u}}^\top \mathcal{V}'_{\bm{xx}} \bfun_{\bm{x}}^\top + \underbrace{\mathcal{V}'_{\bm{x}} \cdot \bfun_{\bm{ux}}}_{\text{DDP terms}}.
\end{align}

The last set of terms in \eref{eq:ddp_backward_pass} is included in DDP, which is a second-order approach, but \textit{excluded} in iLQR, which we described as 1.5-order, as it includes second-order cost derivatives, but not second-order dynamics derivatives.

We then minimize this quadratic approximation w.r.t. the control $\delu$:
\begin{equation}
    \delu^* = \argmin_{\delu} Q(\cdot) = -\bm{Q}_{\bu\bu}^{-1} \bm{Q}_{\bu} - \bm{Q}_{\bu\bu}^{-1} \bm{Q}_{\bu\bx} \delx.
\end{equation}
This defines feed-forward terms $\bm{k}=\bm{Q}^{-1}_{\bm{uu}}\bm{Q_u}$ and state feedback term $\bm{K}=\bm{Q}^{-1}_{\bm{uu}} \bm{Q}_{\bm{ux}}$ at time $t$. Using this result in the expansion of $Q$, we can obtain a set of equations for the optimal cost to go $\mathcal{V}$:
\begin{equation} \label{eq:value_function_update}
\begin{gathered}
    \mathcal{V} = Q(\bx_t, \bu_t^*) = Q(\bx_t^r, \bu_t^r) - \bm{Q_u}\bm{Q}^{-1}_{\bm{uu}} \bm{Q_u}, \nonumber\\
    \mathcal{V}_{\bm{x}} = \bm{Q_x} - \bm{Q_{xu}} \bm{Q}^{-1}_{\bm{uu}} \bm{Q_u}, \nonumber\\
    \mathcal{V}_{\bm{xx}} = \bm{Q_{xx}} - \bm{Q_{xu}} \bm{Q}^{-1}_{\bm{uu}} \bm{Q_{ux}}. 
\end{gathered}
\end{equation}
This process is repeated for time $T$ to $1$ recursively.
Finally, a forward pass from $1$ to $T - 1$ computes the new reference trajectory using the optimal gains. We also use the full, non-linear system dynamics to ensure full feasibility:
\begin{align} \label{eq:forward_pass}
    \hat{\bx}_1 &= \bar{\bx}_1, \nonumber\\
    \delta\hat{\bu}_t &= \hat{\bu}_t - \bu_t^r 
        = - {\bm{k}}_t - {\bm{K}}_t (\hat{\mathbf{x}}_t - \mathbf{x}_t^r), \nonumber\\
    \hat{\bx}_{t + 1} &= \bfun(\hat{\bx}_t, \hat{\bu}_t),
\end{align}
Note that this reduces to LQR when the dynamics are linear, which is solved in a single iteration. In the next section, we describe how to optimize the parameters $\bm{\theta}$ of the optimal control problem using DDP.

\subsection{Optimizing the Parameters of Optimal Control Problems}
To optimize the parameters $\bm{\theta}$ of the OC problem, we begin by formulating a bi-level optimization problem with an inner loop containing the parameterized OC problem, i.e.:
\begin{align}
    \min_{\bm{\theta}, {X}, {U}} &
        \ J_\textsc{UL}(\bm{\theta}, X, U) \nonumber \\
    \text{s.t.}\ & X, U = \textsc{OC}(\bm{\theta}), 
    \label{eq:codesign_formulation}
\end{align}
where $\textsc{OC}(\bm{\theta})$ is the optimal control minimization problem as defined in \eref{eq:oc_cost}. This problem computes the state and control trajectories $X,U$ and can be solved by both DDP and iLQR.

The bi-level optimization minimizes an upper-level cost $J_\textsc{UL}$ by computing a set of optimal parameters $\boldsymbol{\theta}$ along the optimal trajectories $X,U$.
This upper-level cost can be, for instance, the total energy when optimizing the design or an imitation learning loss function or a system identification cost as in \cite{amos_diffmpc_2018}. We can then optimize $\bm{\theta}$ using gradient descent:
\begin{equation} \label{eq:gd}
    \bm{\theta}_{k + 1} = \bm{\theta}_{k} - \eta \nabla_{\bm{\theta}} J_{\textsc{UL}}
\end{equation}
where $\eta$ is the learning rate, $k$ is the iteration, and $\nabla_{\bm{\theta}} J_{\textsc{UL}}$ is the gradient of the upper level cost-function we compute via DDP \footnote{We use $\dd J_\textsc{UL}/\dd\bm{\theta}$ and $\nabla_{\bm{\theta}} J_{\textsc{UL}}$ interchangeably for the total derivative.}.

\section{Derivatives via Differential Dynamic Programming (DDP)}
To obtain $\nabla_{\bm{\theta}} J_{\textsc{UL}}$ via DDP, we first define the optimality conditions of the nonlinear problem. Then we show the equivalent LQR problem that has the same optimality conditions and use the method of \cite{amos_diffmpc_2018} to differentiate it. We show where we need to include the DDP terms, which is our contribution.
\subsection{Optimality Conditions and Newton Update}
We begin by writing the Lagrangian of the nonlinear problem problem we defined in \eref{eq:oc_cost}:
\begin{align}
    \mathcal{L} = &\sum_{t=1}^{T - 1} \left( \ell(\bx_t, \bu_t)
    + \boldsymbol{\lambda}_t^\top \left( \bfun(\bx_t, \bu_t)  - \bx_{t + 1}\right)  \right) + \nonumber\\
    &h(\bx_T) + \boldsymbol{\lambda}_0^\top (\bx_1 - \bar{\bx}_1),
\end{align}
where $\boldsymbol{\lambda}_t$ are Lagrange multiplier associated with the initial conditions and the dynamics. We then write the optimality and primal feasibility conditions (KKT conditions) as:
\begin{equation}
    \frac{\partial \mathcal{L}}{\partial \{\bx_t, \bu_t\}} = \mathbf{0}, \quad \frac{\partial \mathcal{L}}{\partial \boldsymbol{\lambda}_t} = \mathbf{0},
\end{equation}
which form a system of equations. We then apply the Newton method to the KKT conditions, obtaining the following update rule:
\begin{align} \label{eq:kkt_cond}
    &\underbrace{\begin{bmatrix}
        \ddots & & & & & \\
        & \mathcal{L}_{\bx\bx_t}   & \mathcal{L}_{\bx\bu_t}  & \bfun_{\bx_t}^\top &         &         \\
        & \mathcal{L}_{\bu\bx_t}   & \mathcal{L}_{\bu\bu_t}  & \bfun_{\bu_t}^\top &         &         \\
        & \bfun_{\bx_t}  & \bfun_{\bu_t} &              & -\mathbf{I}      &         \\
        &          &         &          - \mathbf{I} & \ddots\\
    \end{bmatrix}}_{\mathbb{K}}
    \begin{bmatrix}
        \vdots \\
        \delx_t \\
        \delu_t
        \\
        \boldsymbol{\lambda}^+_t
        \\
        \delx_{t + 1}
        \\
        \vdots
    \end{bmatrix}
    = - \begin{bmatrix}
        \vdots \\
        \ell_{\bx_t} \\
        \ell_{\bu_t} \\
        \bar{\bm{g}_t} \\
        \ell_{\bx_{t + 1}} \\
        \vdots
    \end{bmatrix},
\end{align}
where $\mathbb{K}$ is the KKT matrix, $\boldsymbol{\lambda}^+_t = \boldsymbol{\lambda}_t + \delta \boldsymbol{\lambda}_t$ is the updated Lagrange multiplier and $\bar{\bm{g}}_t = \bfun(\bx_t,\bu_t) - \bx_{t + 1}$ defines the gaps in the dynamics\footnote{When using DDP or iLQR this gap is $\bf{0}$ due to the forward pass.}. The Hessian of the Lagrangian is:
\begin{align} \label{eq:kkt_ddp}\nonumber
    \mathcal{L}_{\bm{x}\bm{x}_t} &=
        \boldsymbol{\ell}_{\bm{x}\bm{x}_t} + \boldsymbol{\lambda}_t \cdot 
            \bm{f}_{\bm{x}\bm{x}_t} \\
    \mathcal{L}_{\bm{x}\bm{u}_t} =  \mathcal{L}_{\bm{u}\bm{x}_t}^\top &= 
        \boldsymbol{\ell}_{\bm{x}\bm{u}_t} + \boldsymbol{\lambda}_t \cdot 
            \bm{f}_{\bm{x}\bm{u}_t} \\\nonumber
    \mathcal{L}_{\bm{u}\bm{u}_t} &= 
        \boldsymbol{\ell}_{\bm{u}\bm{u}_t} + \boldsymbol{\lambda}_t \cdot 
            \bm{f}_{\bm{u}\bm{u}_t}
\end{align}
The expressions in \eref{eq:kkt_ddp} resemble the DDP update equation for $Q$. This is not a coincidence, as in fact $\boldsymbol{\lambda} \coloneqq \mathcal{V}_{\bx}$ and DDP is an efficient way of inverting the sparse KKT matrix, which we previously described in~\cite{mastalli2020direct}. 
\subsection{Derivatives of the Optimal Control Problem}
To obtain the analytical derivatives of the upper-level cost, we use the chain rule to split it into (i) the derivatives of the cost w.r.t. the DDP parameters $\boldsymbol{\xi}$ and (ii) the derivative of those parameters w.r.t. $\bm{\theta}_i$, the $i^\text{th}$ component of $\bm{\theta}$:
\begin{align}
    \frac{\dd J_\textsc{UL}}{\dd \bm{\theta}_i} = 
    \frac{\partial J_\textsc{ul}}{\partial \bm{\theta}_i}  + \sum_t \frac{\partial J_\textsc{UL}}{\partial \boldsymbol{\xi}_t} \frac{\dd\boldsymbol{\xi}_t}{\dd\bm{\theta}_i}, \text{where}\ \boldsymbol{\xi}_t &= \{
        \bm{F}_t, \bm{C}_t, \bm{c}_t, \bar{\bfun}_t
    \}
\end{align}
where the parameters $\boldsymbol{\xi}_t$ are defined as:
\begin{align}
    \bm{F}_t &= \begin{bmatrix} \bfun_{\bx_t} \\ \bfun_{\bu_t} \end{bmatrix}
    \quad \bm{F}_0 = \bm{0},
    \quad \bm{C}_t = 
        \begin{bmatrix}
            \mathcal{L}_{{\bx\bx}_t} & \mathcal{L}_{{\bx\bu}_t} \\
            \mathcal{L}_{{\bx\bx}_t} & \mathcal{L}_{{\bu\bu}_t}
        \end{bmatrix}, \nonumber\\
    \bm{c}_t &= 
        \begin{bmatrix}
           \ell_{\bx_t} - \mathcal{L}_{{\bx\bx}_t} \bx^r - \mathcal{L}_{{\bx\bu}_t} \bu^r \\
            \ell_{{\bu}_t} - \mathcal{L}_{{\bu\bu}_t} \bu^r - \mathcal{L}_{{\bu\bx}_t} \bx^r
        \end{bmatrix}, \nonumber\\
    \bar{\bfun}_t &= \bfun(\bx_t^r, \bu_t^r; \bm{\theta}) - \bfun_{{\bx}_t} \bx_t^r - \bfun_{{\bu}_t} \bu_t^r,
    \quad \bar{\bfun}_0 = \bar{\bx}_1,
\end{align}
where $\bm{F}_0$ and $\bar{\bfun}_0$ define the initial conditions of the problem.
We now show how these parameters define an LQR problem, which has the same optimality conditions as the original problem. We can then use the approach of \cite{amos_diffmpc_2018} to differentiate this LQR at the optimum. The LQR problem is obtained via a Taylor expansion\footnote{Here we do not include the terminal cost explicitly. Introduce $\delu_T = \bm{0}$ and set $\ell_T(\bx, \bu; \theta) := h(\bx; \theta)$ to obtain the equivalent problem. This avoids the need for an extra term for the terminal cost and makes the rest of the section easier to follow.}:
\begin{align} \label{eq:ddp_lqr}
    \min_{\delx_t, \delu_t} & \sum_t 
    \frac{1}{2}
        \begin{bmatrix}
            \delx_t \\
            \delu_t
        \end{bmatrix}^\top
            \begin{bmatrix}
                \mathcal{L}_{{\bm{x}\bm{x}}_t} & \mathcal{L}_{{\bm{x}\bm{u}}_t} \\
                \mathcal{L}_{{\bm{x}\bm{u}}_t} & \mathcal{L}_{{\bm{u}\bm{u}}_t}
            \end{bmatrix}
        \begin{bmatrix}
            \delx_t \\
            \delu_t
        \end{bmatrix} + 
        \begin{bmatrix}
            \ell_{{\bx}_t} \\
            \ell_{{\bu}_t}
        \end{bmatrix}^\top
        \begin{bmatrix}
            \delx_t \\
            \delu_t
        \end{bmatrix} \nonumber\\
        \text{s.t.}\ &
            \delx_{t + 1} = \begin{bmatrix}
                \bfun_{{\bx}_t} \\
                \bfun_{{\bu}_t}
            \end{bmatrix}^\top
            \begin{bmatrix}
                \delx_t \\
                \delu_t
        \end{bmatrix},
\end{align}
where, importantly, the second-order DDP terms from \eref{eq:kkt_cond} are included (in $\mathcal{L}_{\bf{x}\bf{x}}, \mathcal{L}_{\bf{x}\bf{u}}$ and $\mathcal{L}_{\bf{u}\bf{u}}$). This problem has the same KKT matrix and gives the same update step for $\delx, \delu$ as applying the Newton method in \eref{eq:kkt_cond}. To use~\cite{amos_diffmpc_2018} we need to expand and collect terms to obtain the equivalent LQR problem in $\bx, \bu$-space:
\begin{align}
        \min_{\bx_t, \bu_t} & \sum_t 
    \frac{1}{2}
        \begin{bmatrix}
            \bx_t \\
            \bu_t
        \end{bmatrix}^\top
           \bm{C}_t
        \begin{bmatrix}
            \bx_t \\
            \bu_t
        \end{bmatrix} + 
        \bm{c}_t
        \begin{bmatrix}
            \bx_t \\
            \bu_t
        \end{bmatrix} \nonumber\\
        \text{s.t.}\ &
            \bx_{t + 1} = \bm{F}_t^\top
            \begin{bmatrix}
                \bx_t \\
                \bu_t
        \end{bmatrix} +\bar{\bfun}_t.
\end{align}
where the matrices $\bm{F}_t, \bm{C}_t, \bm{c}_t, \bar{\bfun}_t$ obtained by expanding $\delx, \delu$ and collecting terms from \eref{eq:ddp_lqr} as defined above.
With this, we can compute the derivative of any upper-level cost function $J_\textsc{ul}$ w.r.t. to these parameters following the approach of \cite{amos_diffmpc_2018} by differentiating through the KKT conditions. We first compute the derivative w.r.t. LQR parameters $\boldsymbol{\xi}$.
To compute this, we replace the linear terms on the right side of \eref{eq:kkt_cond} with the derivatives of the upper-level function (and $0$). This forms another LQR problem, which we need to solve only once as we can re-use the KKT matrix $\mathbb{K}$ from the final backward pass of DDP:
\begin{equation}
    \mathbb{K} \begin{bmatrix}
        \vdots \\
        d_{\bx_t} \\
        d_{\bu_t} \\
        d_{\boldsymbol{\lambda}_t} \\
        d_{\bx_{t + 1}} \\
        \vdots
    \end{bmatrix} =
    \begin{bmatrix}
        \vdots \\
        \nabla_{\bx_t} J_{\textsc{UL}} \\
        \nabla_{\bu_t} J_{\textsc{UL}}\\
        \mathbf{0} \\
        \nabla_{\bx_{t + 1}} J_{\textsc{UL}} \\
        \vdots
    \end{bmatrix}
\end{equation}
This linear problem\footnote{This is equivalent to solving an LQR as we discuss in the next section.} gives us the differential terms $d_{\bm{\tau}_t} = \{d_{\bx_t} d_{\bu_t}\}$ and $d_{\boldsymbol{\lambda}_t}$, which are then used to obtain the required derivatives:
\begin{align} \label{eq:derivative_ddp_params}
    \nabla_{\bm{C}_t} J_{\textsc{UL}} &= 
        d_{\boldsymbol{\tau}_t} \otimes \boldsymbol{\tau}_t^* ,\ \ \nabla_{\bm{c}_t} J_{\textsc{UL}} = d_{\boldsymbol{\tau}_t},
    \nabla_{\bar{\bx}_1} J_{\textsc{UL}} = d_{\boldsymbol{\lambda}_{0}}, \nonumber \\
    \nabla_{\bm{F}_t} J_{\textsc{UL}} &= d_{\boldsymbol{\lambda}_{t + 1}} \otimes \boldsymbol{\tau}_t^* + \boldsymbol{\lambda}_{t + 1}^* \otimes d_{\bm{\tau}_t}, 
    \nabla_{\bar{\bfun}_t} J_{\textsc{UL}} = d_{\boldsymbol{\lambda}_{t + 1}},
\end{align}
where $\otimes$ denotes an outer product and $\bm{\tau}_t^* = \{ \bx_t^*, \bu_t^* \}$ and $\boldsymbol{\lambda}_t^*$ are the optimal trajectory and Lagrange multipliers of the original OC problem. Finally, the Lagrange multipliers are obtained by the following recursive set of equations:
\begin{align}
    \boldsymbol{\lambda}_{T}^* &= \boldsymbol{h}_{\bx\bx_t} \bx_T^* + \boldsymbol{h}_{\bx_t}\nonumber\\
    \boldsymbol{\lambda}_t^* &= \bfun_{{\bx}_t}^\top \boldsymbol{\lambda}_{t+1}^* + \boldsymbol{\ell}_{{\bx\bx}_t} \bx_t^* + \boldsymbol{\ell}_{{\bx}_t} + \boldsymbol{\ell}_{{\bx\bu}_t} \bu_t^*
\end{align}
We refer the reader to \cite{amos_diffmpc_2018} for a full derivation of the above result. Next we describe how to include the DDP second-order terms, which is our key contribution.

\subsection{Computing Derivatives via DDP}
This last step involves inverting the KKT matrix in \eref{eq:kkt_cond}. To do so, we can use Riccati recursion too, as it is equivalent to solve a single LQR problem. However, when we do so, we need to include the DDP terms from \eref{eq:kkt_ddp}. We need to solve the following sub-problems at each timestep:
\begin{equation} \label{eq:subproblem_lqr}
    d_{\bu_t} = 
     \argmin_{d_{\bu_t}} \frac{1}{2}
        d_{\bm{\tau}_t}^\top
        \begin{bmatrix}
            \bm{Q}_{\bm{xx_t}} & \bm{Q}_{\bm{xu_t}} \\
            \bm{Q}_{\bm{ux_t}} & \bm{Q}_{\bm{uu_t}}
        \end{bmatrix}
        d_{\bm{\tau}_t}
        + 
        d_{\bm{\tau}_t}^\top
        \begin{bmatrix}
            \bm{Q}_{\bm{x}_t}' \\
            \bm{Q}_{\bm{u}_t}'
        \end{bmatrix}.
\end{equation}
We can re-use the derivatives of the value function from the final DDP iteration:
\begin{align} \label{eq:second_order_q}
    \bm{Q_{xx}} &= \bm{\ell_{xx}} + \bfun_{\bm{x}}^\top \mathcal{V}'_{\bm{xx}} \bfun_{\bm{x}} + \underbrace{\mathrm{V}'_{\bm{x}} \cdot \bfun_{\bm{xx}}} \nonumber \\
    \bm{Q_{xu}} &= \bm{\ell_{xu}} + \bfun_{\bm{x}}^\top \mathcal{V}'_{\bm{xx}} \bfun_{\bm{u}} + \underbrace{\mathrm{V}'_{\bm{x}} \cdot \bfun_{\bm{xu}}} \nonumber \\
    \bm{Q_{uu}} &= \bm{\ell_{uu}} + \bfun_{\bm{u}}^\top \mathcal{V}'_{\bm{xx}} \bfun_{\bm{u}} + \underbrace{\mathrm{V}'_{\bm{x}} \cdot \bfun_{\bm{uu}}}_{\text{DDP terms}},
\end{align}
where $\mathcal{V}'$ is the value function for the LQR and $\mathrm{V}'$ is the value function from the last iteration of the optimal control solver. If using iLQR to solve the OC problem, these DDP terms need to be included to obtain the correct derivatives.
Instead when we use DDP to solve the OC problem, the second-order derivatives of the $\bm{Q}$-function can be directly re-used. To see this, note that they only depend on the second-order derivatives of the value function ($\mathcal{V}_{\bx\bx}'$ and $\mathrm{V}_{\bx\bx}'$). However, those themselves are updated using only second-order derivatives of the $\bm{Q}$-function (\eref{eq:value_function_update}) which are shared between the LQR and DDP (\eref{eq:subproblem_lqr}). 

Finally, the first-order derivatives of $\bm{Q}'$ are different for the LQR and are given by:
\begin{align}
    \bm{Q}_{\bm{x}}' &= \nabla_{\bm{x}} J_{\textsc{UL}} + \bfun_{\bm{x}}^\top \mathcal{V}_{\bx}' \nonumber \\
    \bm{Q}_{\bm{u}}' &= \nabla_{\bm{u}} J_{\textsc{UL}} + \bfun_{\bm{u}}^\top \mathcal{V}_{\bx}'.
\end{align}

Using an LQR to take the derivative of the OC problem without the DDP terms is equivalent to differentiating iLQR as done in \cite{amos_diffmpc_2018}. This can give the incorrect gradient, which as we show experimentally later, may or may not impact the resulting optimization.

So far we have described how to take the derivative of the upper-level cost w.r.t. $\boldsymbol{\xi}$, the parameters of DDP itself. The key was the inclusion of the second-order DDP terms in the Lagrangian of the problem. Next we show how to compute the derivative w.r.t. optimization parameters $\bm{\theta}$.

\subsection{Chain Rule  }
Next, once we have the derivatives of the upper-level cost w.r.t. $\boldsymbol{\xi}$, we use the chain rule to compute the required derivatives w.r.t. $\bm{\theta}_i$ (giving derivatives element-wise to avoid complex tensor notation):
\begin{equation} \label{eq:chain_rule}
    \nabla_{\bm{\theta}_i} J_\textsc{UL} =
    \frac{\dd J_\textsc{UL}}{\dd \bm{\theta}_i} =  
    \frac{\partial J_\textsc{ul}}{\partial \bm{\theta}_i} + \sum_t \frac{\partial J_\textsc{UL}}{\partial \boldsymbol{\xi}_t} \frac{\dd\boldsymbol{\xi}_t}{\dd\bm{\theta}_i}.
\end{equation}
The derivative term $\frac{d\xi}{d\bm{\theta}_i}$ also requires second-order derivative terms w.r.t dynamics. The authors in \cite{amos_diffmpc_2018} use automatic differentiation and \textsc{Pytorch}~\cite{paszke2019pytorch} for this derivative. For robotics problems, however, automatic differentiation is not always available, especially when using rigid body dynamics libraries (e.g., \textsc{Crocoddyl}~\cite{mastalli_crocoddyl_2020}) and computing this last step needs to be done analytically (up to the dynamics themselves). Indeed one must be careful when deriving them, as a recursive update is required. To make our discussion complete, we give these derivatives analytically. To obtain the derivative we firstly expand the chain rule:
\begin{align}
    \frac{\dd J_\textsc{UL}}{\dd \bm{\theta}_i} &=  \frac{\partial J_\textsc{ul}}{\partial \bm{\theta}_i} +
        \sum_t  
        \left(\frac{\partial J_\textsc{UL}}{\partial {\bm{F}_t}}\right)^T \frac{\dd \bm{F}_t}{\dd \bm{\theta}_i} +
        \left(\frac{\partial J_\textsc{UL}}{\partial {\bar{\bfun}_t}}\right)^T \frac{\dd \bar{\bfun}_t}{\dd \bm{\theta}_i} \nonumber\\
        &+ 
        \left(\frac{\partial J_\textsc{UL}}{\partial {\bm{C}_t}}\right)^T
            \frac{\dd \bm{C}_t}{\dd \bm{\theta}_i} +
        \left(\frac{\partial J_\textsc{UL}}{\partial {\bm{c}_t}}\right)^T
            \frac{\dd \bm{c}_t}{\dd \bm{\theta}_i}.
\end{align}
We then simply plug in the expressions from \eref{eq:derivative_ddp_params} into the expanded chain rule, obtaining:
\begin{align} \label{eq:chain_rule}
    \frac{\dd J_\textsc{UL}}{\dd \bm{\theta}_i} &= 
        \frac{\partial J_\textsc{ul}}{\partial \bm{\theta}_i}
            + \sum_t \bigg(
         \bigg\langle\bm{\lambda}_t^* \otimes d_{\bm{\tau}_t}, 
         \begin{bmatrix}
            \nabla_{\bm{\theta}_i} \bfun_{\bx_t} \\
            \nabla_{\bm{\theta}_i} \bfun_{\bu_t}
         \end{bmatrix} \bigg\rangle_{\text{F}} \nonumber\\
    &\ +
        \langle d_{\bm{\lambda}_t}, \bfun_{\bm{\theta}_i} \rangle_{\text{F}} +
        \bigg\langle d_{\bm{\tau}_t},
            \begin{bmatrix}
                \ell_{\bx_t \bm{\theta}_i} \\
                \ell_{\bu_t \bm{\theta}_i}
            \end{bmatrix}  \bigg\rangle_{\text{F}} \bigg),
\end{align}
where $\otimes$ is the outer product, $\cdot$ denotes a tensor dot product (contraction), and $\langle\, ,\, \rangle_{\text{F}}$ denotes the Frobenius product, which is an element-wise product followed by a summation\footnote{This notation is unusual, but required, since we differentiate w.r.t. matrices, which means w.r.t. each matrix element. This is why we need element-wise operations for each matrix element.}. To obtain the derivatives of $\bfun_{\bx}$ and $\bfun_{\bu}$ w.r.t. $\bm{\theta}_i$ we use the following recursive expression:
\begin{align}
    \nabla_{\bm{\theta}_i} \bfun_{\bx_t} &= \bfun_{\bx\bm{\theta}_i} + \bfun_{\bm{xx}_t} \cdot \frac{\dd \bx_t}{\dd \bm{\theta}_i}, \nonumber\\
    \nabla_{\bm{\theta}_i} \bfun_{\bu_t} &= \bfun_{\bu\bm{\theta}_i} + \bfun_{\bm{ux}_t} \cdot \frac{\dd \bx_t}{\dd \bm{\theta}_i}, \nonumber\\
    \frac{\dd \bx_t}{\dd \bm{\theta}_i} &= \bfun_{\bm{\theta}_i} + \bfun_{\bx_t} \frac{\dd \bx_{t - 1}}{\dd \bm{\theta}_i},
\end{align}
where the last line follows directly from the nonlinear rollout performed in the forward pass (\eref{eq:forward_pass}).
Note specifically that we have no third-order derivatives in this expression, which at first seems necessary as, for instance, we are differentiating $\mathcal{L}_{\bx\bx}$ w.r.t. $\bm{\theta}_i$. This is actually expected (indeed these terms cancel out), as we started by differentiating the first-order optimality conditions. This final expression thus should only have first and second-order derivatives.

\begin{algorithm*} \label{alg:dddp}
    \caption{Analytical Derivatives of Optimal Control via DDP}
    \label{alg:dddp}
    \begin{algorithmic}[1]
    \Procedure{Diff-DDP}{$J_\textsc{UL}(\cdot)$, $\bm{\theta}$}
      \State $X, U = \text{OC}(\bm{\theta})$ \Comment{Compute the optimal trajectory using the solver.}
      \State Run an LQR to compute $\partial J_\textsc{UL}/\partial \{ \bm{F}_t, \bm{C}_t, \bm{c}_t, \bar{\bfun}_t \}$, adding second order DDP terms to the KKT matrix $\mathbb{K}$ (\eref{eq:second_order_q}).
      \State Use the chain rule to obtain $\dd J_\textsc{UL}/\dd\bm{\theta}_i$ (\eref{eq:chain_rule}) for every element $\bm{\theta}_i$.
    \EndProcedure
    \end{algorithmic}
\end{algorithm*}

\subsection{Summary of Algorithm}
Our algorithm is summarized in Alg. \autoref{alg:dddp}. For solving the optimal control problem it does not matter whether first or second-order terms are used, so long as the optimal control solver converges. We can thus only include the second order terms only when computing derivatives (\eref{eq:second_order_q}) making this process efficient. %

\section{Experiments}
To experimentally show the need for DDP derivatives (versus iLQR derivatives), we studied three systems: a pendulum, a double pendulum and the Kinova robotic arm. We started by validating the accuracy of our derivatives on the pendulum and double pendulum systems.

\subsection{Validation of Gradients}
For each experiment, we define an upper-level optimization vector $\bm{\theta} = \{
    \bm{\rho},
    q_f
\}$,
where $\bm{\rho}$ defines the parametric dynamics used in each of the systems (described below in each subsection) and $q_f$ defines the diagonal entries of the weighting term at the terminal state cost:
\begin{align}
    h(\bx_T; \bm{\theta}) &= (\bx_T - \bx^*)^T \underbrace{
        \textsc{diag}(q_f, \dots, q_f)
    }_{\bm{Q}_f} (\bx_T - \bx^*)
\end{align}
where $\textsc{diag}$ creates a diagonal matrix and $\bx^*$ is the goal state.
For the pendulum and double pendulum, we first sampled $100$ points for the vector $\bm{\theta}$ with problem-specific bounds defined below. Then we compared the average gradient error versus  \textit{Automatic Differentiation} (AD), which unrolls the entire optimization loop. We used the \textsc{ForwardDiff}~\cite{forwarddiffjulia} package in \textsc{Julia}~\cite{Julia} for automatic differentiation. We computed the error as the absolute difference:
\begin{equation}
    \mathcal{G}_\textsc{ERR} = \sum_{\bm{\theta}_i} |\nabla_{\bm{\theta}_i}^\textsc{AD}J_\textsc{UL} - \nabla_{\bm{\theta}_i}^\textsc{SA}J_\textsc{UL}|,
\end{equation}
where, again, $\textsc{SA}$ stands for sensitivity analysis derivatives. We computed this for both DDP derivatives (our approach) as well as iLQR derivatives~\cite{amos_diffmpc_2018}.

Additionally, we report results for both $64$-bit and $128$-bit float precision. The optimality conditions for any sensitivity-based approach hold only if the solver has converged. Since we are looking at accumulating gradients over a relatively large number of timesteps ($T=50$ timesteps) convergence with better precision allowed us to evaluate whether any errors we observe are due to missing terms or simply due to poor convergence.

We used the common gradient convergence metric $\bm{Q_u} \bm{Q}_{\bm{{uu}}}^{-1} \bm{Q_u}$ and specify a threshold of convergence of $10^{-15}$ when using $64$-bit precision and $10^{-30}$ when using $128$-bit precision. We re-sampled if the motion planning did not converge for $64$-bit precision. For $128$-bit precision we used the same samples as for $64$-bit precision. Of those  $0$ and $4$ samples did not converge when using $128$-bit precision for the pendulum and double pendulum, respectively. Since few samples were excluded (less than $5$ percent), we did not resample when using $128$-bit precision.
Next we describe our results on the pendulum and double pendulum systems.

\subsubsection{Pendulum (SysID/Imitation Learning)}
The first system we studied is a simple pendulum (e.g. \cite[Chapter~2]{underactuated}). The pendulum is attached at the origin and has a parametrized link length $\bm{\rho}$. The state space is comprised of the angle and angular velocity of the joint and the controls are the torques applied at the joint. Thus the dimensions of the state space are $N_x=2$ and $N_u=1$. %

We generated a swing-up trajectory for the pendulum with a known link length of $\rho = \SI{0.5}{\meter}$, $q_f = 10^3$. The trajectory has $T = 50$ knots with a timestep of $\Delta t = 10^{-2}$ seconds for a trajectory that lasts half a second. We defined an imitation learning/system identification upper level cost in the same way as the authors in \cite{amos_diffmpc_2018} did:
\begin{equation}
    J_{\textsc{UL}} = \sum_{t = 1}^{T=49} ||\bu_t - \bu_t^i||_2,
\end{equation}
where $\bu_t^i$ is the imitation learning trajectory generated above. The goal of optimizing this function is to find the pendulum link length (system identification) together with the $q_f$ (imitation learning).
Note that this is one important benefit of differentiable optimal control, as we can solve both problems jointly.
We then computed the gradient of the upper-level function w.r.t. the link length $\bm{\theta} = \{\rho,q_f\}$ for link lengths in the range $\rho \in [0.1, 1.0]\SI{}{\meter}$ and $q_f \in [1, 10^{4}]$. We additionally included the known optimum ($\rho = \SI{0.5}{\meter}$, $q_f = 10^3$) when computing the errors.

\begin{figure}[t]
    \centering
    \includegraphics[width=0.49 \textwidth]{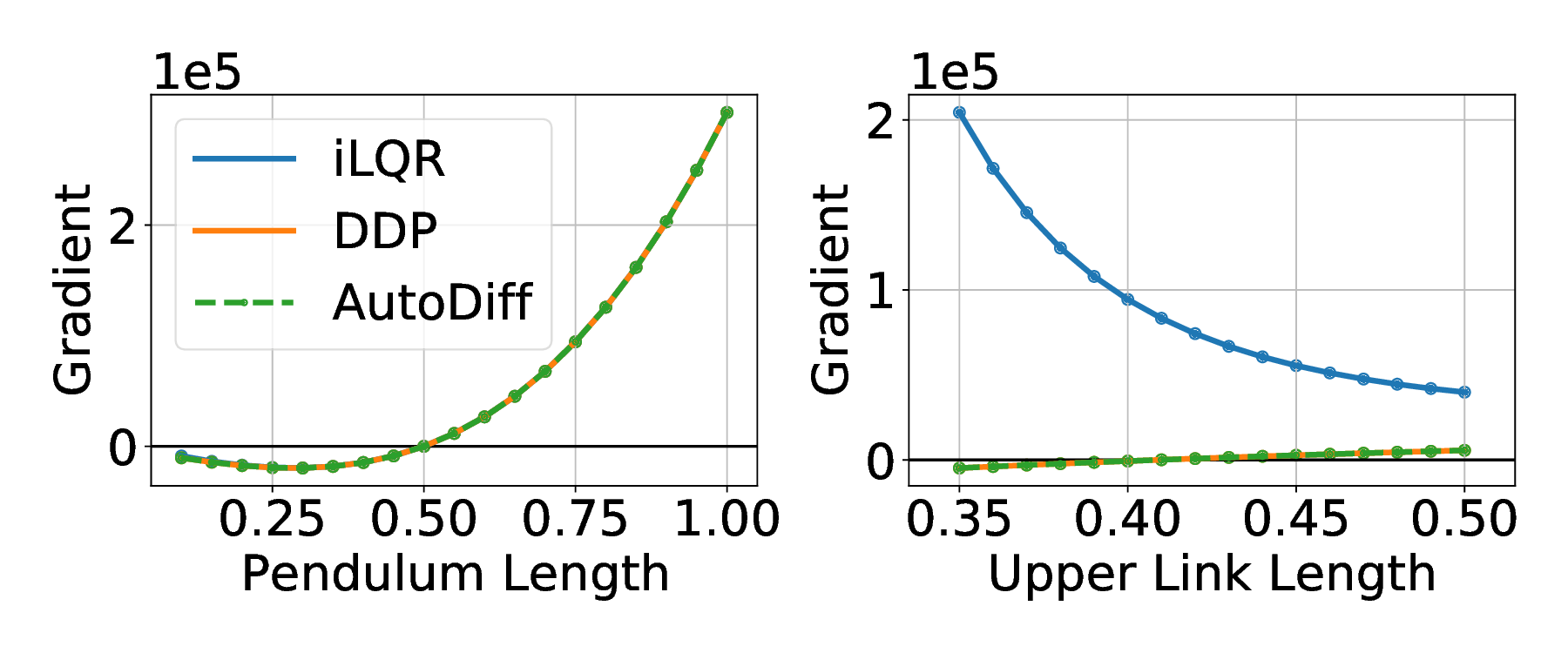}
    \caption{Pendulum (left) and double pendulum (right). For the pendulum all methods give the same optimum and sign, however, for the double pendulum iLQR derivatives have the wrong magnitude and sign.}
    \label{fig:pendulum_gradients}
\end{figure}

The numerical results for all variables are shown in \autoref{tab:gradients} (Pendulum). We can see that using DDP terms leads to least error and the error reduces significantly when using $128$-bit precision. However, for this problem we also see that the minimum error for all methods is small for $64$-bit precision and goes to $0$ for $128$-bit precision. This is in fact at the known optimum for which both methods correctly give a gradient of $0$. Furthermore, the sign of the gradient is correct for both methods everywhere. This suggests that the errors when using iLQR gradients are only in the gradient magnitudes. This would suggest that for optimization purposes it does not matter which derivative computation method we choose.

To illustrate this point further, we fixed $q_f$ to the optimum and plot the gradient w.r.t. the link length of the pendulum in \autoref{fig:pendulum_gradients} (left). Although there are errors for the gradients away from the optimum we cannot see them visually and all gradients intersect with the origin at the same point (the optimal $\rho = \SI{0.5}{\meter}$).

The fact that all methods give the same optimum is an important observation that comes from the upper-level cost gradients. We remind the reader that to compute them we ran another LQR pass with the following substitutions:
\begin{equation}
    \bm{c}_t = \begin{bmatrix}
        \frac{\partial J_{\textsc{UL}}}{\partial \bx_t},
        \frac{\partial J_{\textsc{UL}}}{\partial \bu_t}
    \end{bmatrix}^T,\ 
    \bar{\bfun}_t = \bm{0}.
\end{equation}
If the partials above are $\bm{0}$ when the derivative of interest $\nabla_{\bm{\theta}} J_{\textsc{UL}}$ is also $\bm{0}$, then all methods will give a gradient of $\bm{0}$ at the same value of $\bm{\theta}.$\footnote{Note that this error is non-zero for 64-bit precision in \autoref{tab:gradients}. This is due to that algorithmic differentiation gives differently wrong gradients for $64$-bit precision, again due to convergence errors. This is another reason why we need $128$-bit precision for comparison.} Note in this case the upper-level function only has gradients w.r.t. $\bu_t$ and is $\bm{0}$ when the controls are the same as the imitation learning baseline.

In conclusion, for this task the gradient signs are correct for both iLQR and DDP and both are usable in optimization.

\begin{table*}[t]
    \vspace{4pt}
    \centering
    \begin{tabular}{llcccccccc}\toprule
    & & \multicolumn{4}{c}{Pendulum} & \multicolumn{4}{c}{Double Pendulum}
    \\\cmidrule(lr){3-6}\cmidrule(lr){7-10}
               & & Min err.  & Max err. & Mean err. & Sign errors & Min err.  & Max err. & Mean err. & Sign errors   \\\midrule
    64 bit & iLQR derivative
        & 1.65e-05 & 1.63e+03 & 2.08e+02 & 0/100 %
        & 1.42e+04 & 5.45e+05 & 1.04e+05 & 18/100 \\ %
    & DDP derivative (ours)
        & \textbf{4.59e-07} & \textbf{9.61e-03} & \textbf{4.11e-04} & 0/100
        & \textbf{2.41e-04} & \textbf{1.60e-01} & \textbf{2.13e-02} & \textbf{0/100}
    \\\midrule
    128 bit & iLQR derivative
        & 0 & 1.63e+03 & 2.08e+02 & 0/100
        & 1.42e+04 & 4.44e+05 & 8.78e+04 & 18/100 \\
    & DDP derivative (ours)           
        & 0 & \textbf{5.76e-14} & \textbf{4.77e-15}  & 0/100
        & \textbf{1.42e-15} & \textbf{4.05e-10} & \textbf{4.53e-12} & \textbf{0/100}
    \\\bottomrule
    \end{tabular}
    \caption{Gradient errors as compared to Automatic Differentiation for iLQR and DDP gradients. Using iLQR derivatives gives large errors and on the double pendulum, can give wrong signs. With 128-bit precision the solver can achieve better numerical convergence -- DDP gradients achieve significantly lower error (maximum below $1\mathrm{e}{-9}$) on both problems.}
    \label{tab:gradients}
\end{table*}

\subsubsection{Double Pendulum (SysID/Imitation Learning)}

In this section we will show an example of an upper level cost function and system for which the gradient sign and magnitude are wrong when using iLQR. 

To begin, we define the double pendulum system (e.g. \cite[Appendix~B]{underactuated}). It consists of two links with a point mass at the end of each link. Each of the two link lengths is parameterized as $\bm{\rho} = \{ l_1, l_2 \}$. The state space consists of the angle and angular velocity of each link and the controls are the torques applied at the joints. Thus $N_x = 4$ and $N_u = 2$. The goal is to swing the pendulum from the bottom to the top position in $T = 50$ knots of $\Delta t = 10^{-2}$ seconds for a total trajectory length of half a second.

We generated a trajectory with $\bm{\rho} = \{ \SI{0.5}{\meter}, \SI{0.5}{\meter} \}$, $q_f = 10^{3}$. The imitation learning trajectory is different, however:
\begin{equation}
    J_{\textsc{UL}} =  \dot{\bm{q}}_{50}^T
        \dot{\bm{q}}_{50}  +
        \sum_{t = 1}^{49} ||\bu_t - \bu_t^i||_2 + \dot{\bm{q}}_t^T \dot{\bm{q}}_t, 
\end{equation}
where we added a velocity cost on $\dot{\bm{q}}$, the angular velocity of the double pendulum. This cost was used for imitation learning/SysID of the target trajectory with the added objective of minimizing the joint velocity of the robot as well.

The resulting gradient errors for this cost and samples in the range $l_1, l_2 \in [0.25, 0.5]\si{\meter}$ and $q_f \in [10^2, 10^4]$ are in \autoref{tab:gradients} (Double Pendulum). Once again, using DDP terms leads to the least amount of error.

More importantly, the iLQR derivatives have errors in sign as well ($18$ out of the $100$ samples had the wrong sign). To illustrate this point further, we plot the gradient for the second link length while holding all other variables constant in \autoref{fig:pendulum_gradients} (right). Indeed we see that the iLQR gradients are always positive. Autodiff (AD) and the DDP gradients overlap in this figure. We thus conclude that on this system it is important to use DDP terms when computing gradients.

\subsection{Co-Design with a Kinova Manipulator}
In this experiment we scaled up our approach to a robotic manipulator that demonstrates a real-world use-case for differentiable OC. We used \textsc{Crocoddyl}~\cite{mastalli_crocoddyl_2020} as our motion planning library and show how to make it differentiable for the Kinova manipulator introduced in ~\cite{campeau2019kinova}.

\begin{figure}[t]
    \centering
    \includegraphics[width=0.45 \textwidth]{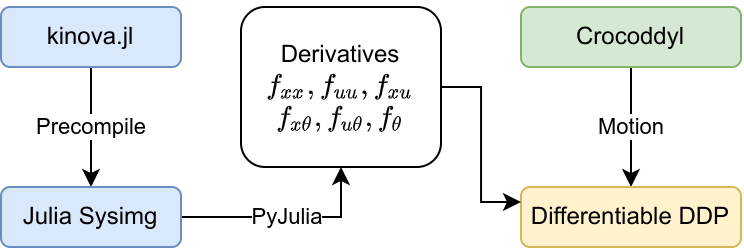}
    \caption{Pipeline for computing second-order derivatives of the rigid-body forward dynamics of the Kinova arm. These terms are needed for obtaining the analytical derivatives of the optimal control problem via DDP using \textsc{Crocoddyl}~\cite{mastalli_crocoddyl_2020}}
    \label{fig:kinova_pipeline}
\end{figure}

We parameterized three of the links of the Kinova 7-DOF manipulator as follows: $\bm{\theta} = \{l_1, l_2, l_3\}$. These three links are represented as cylinders (as shown in \autoref{fig:poster}) and the link inertias are computed based on the link lengths.

We formulated a motion planning problem that aims to reach the goal end-effector position of $[0.5, -0.5, 0.5] \SI{}{\meter}$. The initial robot configuration, the end effector trajectory, and the goal are also shown in \autoref{fig:poster}. The motion planning horizon is specified by $T=50$ knots with $\Delta t = 0.02$ seconds for an optimization horizon of $1$ second.
We then defined a co-design problem with an upper-level cost on the velocity of the joints together with a reachability cost that ensures the target is always reachable in the lower level: %
\begin{equation}
    J_{\textsc{UL}}(\bm{q}, \bm{\theta}) = \sum_{t = 1}^{T} \dot{\bm{q}}_t^T \dot{\bm{q}}_t + \underbrace{10^{3}\ \max\left(0, d - \sqrt{l_1^2 + l_2^2 + l_3^2}\right)^2}_{\textsc{Reachability}}
\end{equation}
where $d$ is the distance to the target, adjusted for the fixed-length robot links.

To compute the analytical derivative of the upper-level cost function, we need the second-order dynamics derivatives, as well as the derivatives of the dynamics w.r.t. the link lengths of the robot, which are currently unavailable analytically in rigid body dynamics libraries such as \textsc{Pinocchio}~\cite{carpentier2019pinocchio} used by \textsc{Crocoddyl}, or \textsc{RBD.jl}~\cite{koolen2019julia}, which we used for previous experiments in this work.

\textsc{RBD.jl}, however, supports algorithmic differentiation. Therefore, we created a robot model programmatically and inserted dual variables for the link lengths (to compute the required gradients). We only need to do this once after the optimization completes (as shown in Algorithm \autoref{alg:dddp}). We then re-used the factorization of the KKT matrix computed by \textsc{Crocoddyl} in order to compute the upper level cost. An illustration of our pipeline is in \autoref{fig:kinova_pipeline}.

We optimized the robot designs using gradient descent (\eref{eq:gd} with $\eta = 5\times10^{-4}$). This co-design problem optimized the robot design that minimizes the joint velocity for the task of reaching a given end-effector position.
\begin{figure}[t]
    \centering
    \includegraphics[width=0.29 \textwidth]{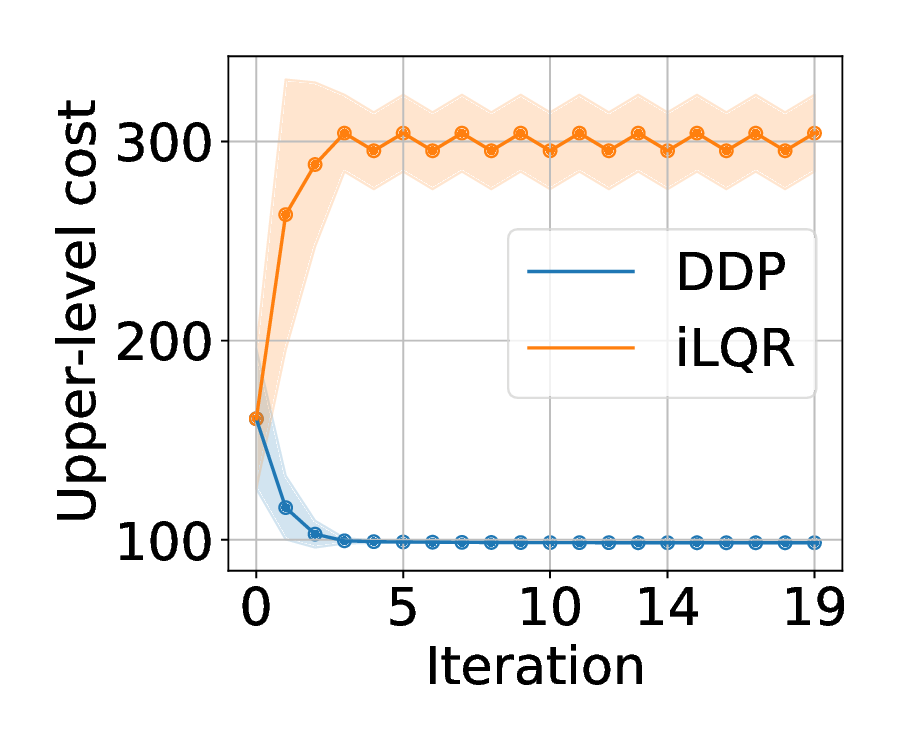}
    \caption{Kinova co-design optimization. Mean cost (joint velocity) and one standard deviation is shown. Using iLQR derivatives leads to wrong gradients that lead to divergence.}
    \label{fig:kinova_optimization}
\end{figure}
To evaluate both iLQR and DDP derivatives more robustly, we sampled and optimized $50$ initial robot designs (link-lengths) uniformly between $\SI{0.1}{\meter}$ and $\SI{0.5}{\meter}$ (for all three links), and also set the limits of the link lengths to $[0.1, 0.5] \SI{}{\meter}$.

\autoref{fig:kinova_optimization} shows the mean costs and standard deviation, and \autoref{fig:poster} shows an example design and the mean optimal designs (as found by our method). The mean optimal design found was $\bm{\theta} = \{ 0.330 \pm \SI{0.003}{\meter}, 0.289 \pm \SI{0.003}{\meter}, 0.1 \pm \SI{0.0}{\meter} \}$ with a maximum standard deviation of $\SI{3}{\milli\meter}$. This design aligns the target with the end-effector on the $z$-plane, leading to minimum velocity. Using DDP derivatives is crucial for this problem, as the gradient descent optimization quickly diverges if we use the iLQR derivatives. Additionally, we observe little variation between samples when using the correct DDP derivatives and the optimization quickly converges (in fewer than 5 iterations) to the optimal design.

\section{Discussion and Future Work}
\label{sec:limitations}
For the Kinova arm we still needed to use algorithmic differentiation (AD) to compute the terms $\bfun_{\bx\bx}$, $\bfun_{\bx\bu}$, $\bfun_{\bu\bu}$ as well as $\bfun_{\bx\bm{\theta}}$, $\bfun_{\bu\bm{\theta}}$ and $\bfun_{\bm{\theta}}$. We used RigidBodyDynamics.jl~\cite{koolen2019julia} (\autoref{fig:kinova_pipeline}) and ForwardDiff.jl~\cite{forwarddiffjulia} to compute these as they are not readily available analytically in rigid body dynamics libraries. This results in the backwards pass having significant computational cost ($187.403 \pm 81.453 \SI{}{\second}$), which is still nonetheless within the same order of magnitude as iLQR at $179.758 \pm 73.462\SI{}{\second}$. This is prohibitive for imitation learning and online SysID, but for co-design computational speed is less important than accuracy. 

In fact recently in the robotics community there have been results showing the benefits of using second-order dynamics in DDP to solve OC problems. This involves the terms $\bfun_{\bx\bx}$, $\bfun_{\bx\bu}$, $\bfun_{\bu\bu}$, and an efficient way to compute the tensor product in \eref{eq:ddp_backward_pass} ~\cite{singh2022efficient, singh2022closed}. A direction for future work is the use of the efficient algorithms developed in \cite{singh2022closed}. Moreover, we still lack efficient algorithms for computing derivatives w.r.t. design variables including second-order derivatives and their tensor products in \eref{eq:chain_rule} -- $\bfun_{\bx\bm{\theta}}$, $\bfun_{\bu\bm{\theta}}$ and $\bfun_{\bm{\theta}}$. Future work lies in developing such algorithms in order to enable online deployment of differentiable solvers for learning and system identification on real robots.

Finally, our approach is a counterpart to the approach of \cite{jin2020pontryagin}. Whereas our results use Value-iteration to compute derivatives, the authors in \cite{jin2020pontryagin} use the Hamiltonian and Pontryagin's Maximum Principle. These two approaches are in theory mathematically equivalent. In practice, our approach is preferred when the solver itself is iLQR or DDP, as the factorization can be re-used rather than having to compute the Hamiltonian. As both approaches involve solving a single LQR to compute derivatives, the complexity is the same. 

\bibliographystyle{IEEEtran}
\bibliography{IEEEfull,IEEEconf,codesign,manual}

\end{document}